\relax
\documentclass[runningheads]{llncs}
\usepackage{times}  % DO NOT CHANGE THIS
\usepackage{helvet} % DO NOT CHANGE THIS
\usepackage{courier}  % DO NOT CHANGE THIS
\usepackage[hyphens]{url}  % DO NOT CHANGE THIS
\usepackage{graphicx} % DO NOT CHANGE THIS
\graphicspath{ {./images/} }
\usepackage[normalem]{ulem}
\usepackage{acronym}
\usepackage{xspace}
\usepackage{paralist} %inline enumeration
\usepackage[title]{appendix}
\usepackage{dingbat}
\usepackage[normalem]{ulem}
\useunder{\uline}{\ul}{}

\usepackage{amssymb}
\usepackage{amsmath}
\DeclareMathAlphabet{\mathpzc}{OT1}{pzc}{m}{it}
\usepackage{multirow}
\usepackage{listings}
\usepackage{tabularx}
\usepackage{float}
\usepackage{tikz-cd}
\usepackage{ bbold }
\usepackage{pgfplots}
\usepackage{hhline}
\usepackage{xcolor}

\title{PaRoT: A Practical Framework for Robust Deep Neural Network Training}

\author{Edward W. Ayers\inst{1},
 Francisco Eiras\inst{2}, Majd Hawasly\inst{2}, Iain Whiteside\inst{2}}

\authorrunning{E.W. Ayers, F. Eiras, M. Hawasly, I. Whiteside}

\institute{DPMMS, Cambridge University, United Kingdom, 
\email{e.w.ayers@maths.cam.ac.uk}
\and
FiveAI, 20 Cambridge Place, Cambridge, United Kingdom, \email{\emph{first.last}@five.ai}
}

\newcommand{\real}{\mathbb{R}}
\renewcommand{\o}{\circ}
\newcommand{\F}{\mathcal{F}}
\newcommand{\D}{\mathcal{D}}
\newcommand{\B}{\mathpzc{B}}
\renewcommand{\P}[1]{\mathcal{P}(#1)}
\renewcommand{\c}{\mathbf{c}}
\renewcommand{\b}{\mathbf{b}}
\newcommand{\s}{\mathbf{s}}

\renewcommand{\j}{\mathbf{t}_j}
\newcommand{\x}{\mathbf{x}}
\newcommand{\y}{\mathbf{y}}
\newcommand{\z}{\mathbf{z}}
\newcommand{\E}{\mathbf{E}}
\renewcommand{\S}{\mathbf{S}}
\newcommand{\T}{\mathbf{T}}
\renewcommand{\v}{\mathbf{v}}
\newcommand{\w}{\mathbf{w}}

\newcommand{\abs}[1]{\left|#1\right|}
\newcommand{\norm}[2]{\lVert{#2}\rVert_{#1}}
\renewcommand{\L}{\mathcal{L}}
\DeclareMathOperator*{\amax}{\mathrm{argmax}}
\DeclareMathOperator*{\amin}{\mathrm{argmin}}
\newcommand{\argmax}[1]{\underset{#1}{\amax}}
\newcommand{\argmin}[1]{\underset{#1}{\amin}}
\newcommand{\Tensor}{tensor\xspace}
\newcommand{\Op}{op\xspace}
\newcommand{\framework}{\emph{PaRoT}\xspace}

\lstdefinestyle{lst_style}{
    basicstyle=\ttfamily\footnotesize,
    breakatwhitespace=false,
    breaklines=true,
    captionpos=b,
    keepspaces=true,
    numbers=left,
    numbersep=4pt,
    showspaces=false,
    showstringspaces=false,
    showtabs=false,
    tabsize=2
}

\usepackage{lstautogobble}  % Fix relative indenting
\usepackage{color}          % Code coloring
\usepackage{zi4}            % Nice font

\definecolor{bluekeywords}{rgb}{0.13, 0.13, 1}
\definecolor{greencomments}{rgb}{0, 0.5, 0}
\definecolor{redstrings}{rgb}{0.9, 0, 0}
\definecolor{graynumbers}{rgb}{0.5, 0.5, 0.5}

\usepackage[linesnumbered,ruled,vlined]{algorithm2e}

\SetCommentSty{mycommfont}

\usepackage{listings}
\newcommand{\Z}{Z}
\acrodef{dnn}[DNN]{Deep Neural Network}
\acrodef{av}[AV]{Autonomous Vehicle}
\acrodef{smt}[SMT]{Satisfiability Modulo Theories}
\acrodef{pgd}[PGD]{Projected Gradient Descent}
\acrodef{fgsm}[FGSM]{Fast Gradient Sign Method}
\acrodef{cnn}[CNN]{Convolutional Neural Network}
\newcommand{\etal}{\emph{et al.}\xspace}

\newcommand{\eg}{e.g.,\xspace}
\newcommand{\ie}{i.e.,\xspace}

\newcommand{\five}{Five AI\xspace}
\newcommand{\bx}{\textsc{Box}\xspace}
\newcommand{\zono}{\textsc{Zonotope}\xspace}
\newcommand{\hz}{\textsc{HybridZonotope}\xspace}
\newcommand{\ox}{\overline{x}}
\newcommand{\ux}{\underline{x}}
\usepackage[caption=false]{subfig}

\begin{document}

\maketitle

\begin{abstract}

\acp{dnn} are finding important applications in safety-critical
systems such as \acp{av}, where perceiving the environment 
correctly and robustly is necessary for safe operation.
Raising unique challenges for assurance due to their
black-box nature, \acp{dnn} pose a fundamental 
problem for regulatory acceptance of these types of systems.
Robust training --- training to minimize excessive sensitivity to small changes in input --- has emerged as one 
promising technique to address this challenge.
However, existing robust training tools are 
inconvenient to use or apply to existing codebases 
and models: they typically only support a small subset of model
elements and require users to extensively rewrite the training code.
In this paper we introduce a novel framework, \framework,
developed on the popular TensorFlow platform, that 
greatly reduces the barrier to entry. Our framework 
enables robust training to be performed on existing
\acp{dnn} without rewrites to the model.
We demonstrate that our framework's performance is comparable 
to prior art, and exemplify its ease of use on 
off-the-shelf, trained models and its testing capabilities on a real-world industrial application: a traffic light detection network.

\end{abstract} 
\acresetall
\section{Introduction} 
\acp{dnn} are finding important applications in safety-critical
systems, such as \acp{av}, where perceiving a complex environment
correctly and robustly is necessary for safe
operation~\cite{AVSurvey,DeepLearning,EndToEnd}. 
The challenge of assuring these so-called \emph{AI-enabled 
systems} is well-known~\cite{Koopman} and has attracted the attention
of researchers and research bodies, \eg DARPA~\cite{assuredautonomy}.
Existing standards and techniques --- such as the
ubiquitous `V' model
--- lean heavily on the existence of a clear
specification to verify against~\cite{safe-ml}. 
Unfortunately, the very nature of deep learning --- where the
specification is implicit in the training data --- poses a fundamental
problem for regulatory acceptance of these systems in a 
safety-critical domain.

\begin{figure}[h]
    \centering
    \subfloat[\label{fig:normal-and-adv-1}]{%
      \includegraphics[width=0.21\textwidth]{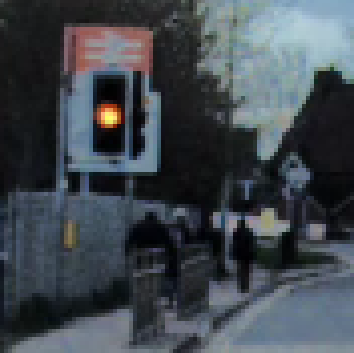}%
    }\hspace{2em}
    \subfloat[\label{fig:normal-and-adv-2}]{%
      \includegraphics[width=0.21\textwidth]{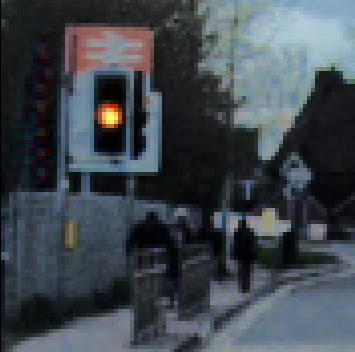}%
    }\hspace{2em}
    \subfloat[\label{fig:normal-and-adv-3}]{%
      \includegraphics[width=0.21\textwidth]{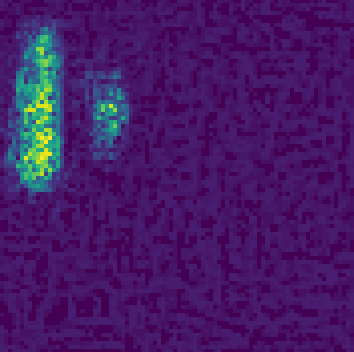}%
    } \hfill
    \subfloat[\label{fig:adv-detection}]{%
      \includegraphics[width=0.53\textwidth]{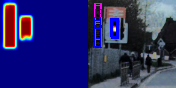}%
    }
    \caption{\textit{Traffic Light Detection Network}: (a) an image 
    from the test set in which the traffic light is identified
    correctly; (b) an adversarial example: a subtly modified version 
    of the original image, identified using \framework; (c) norm of 
    the difference between the original image and the adversarial 
    images; (d) the inference result on the adversarial example, 
    with a confidence heatmap on the left and bounding boxes of 
    the identified traffic lights on the right.}
    \label{fig:traffic-light}
\end{figure}

One of the most troubling features of \acp{dnn} is their
`intriguing' susceptibility to \emph{adversarial examples}:
imperceptible perturbations in the input space that cause
a large change in the output space.
For example, causing an object detection network to misclassify an
image~\cite{intriguing}.
Figure~\ref{fig:traffic-light} shows an adversarial 
example on a traffic light detector. 

The formal verification community has responded to this provocation with 
gusto~\cite{marabou,reluplex,corina,DBLP:conf/kr/AkitundeLMP18,safetyverification,rudy-unified}.
Exacerbating the verification challenge is the indirect 
nature of any `fixes' that can be applied to failure of post-hoc formal
verification for a \ac{dnn}: typically an augmentation to the training 
set.
Unlike with traditional software, fixes to \acp{dnn} can feel very much
like playing a game of whack-a-mole.

The emerging \emph{robust training} paradigm, which integrates the
verification process directly into the training scheme, 
is, in our view,
the most promising approach towards formally verified neural networks. 
The goal of robust training is to minimize a so-called
worst-case adversarial loss.
Formally, let $N_{\theta}: \real^p \to \real^q$
be a neural network with $p$ input features and $q$ outputs, 
 parameterized with weights $\theta$ .
Let $\B_{\epsilon}(x)$ be an $\ell_\infty$-ball of radius 
$\epsilon$ around an input point $x \in \real^p$. 
For a given loss function $\L$, we can define the
\emph{worst-case adversarial loss} $\L_{N_{\theta}}$ at a point $x$ as:
\begin{equation}
    \L_{N_{\theta}}(x,y) := 
    \underset{\tilde{x} \in \B_{\epsilon}(x)}{\mathrm{max}} \L(N_{\theta}(\tilde{x}),y)
\end{equation}

In general, one may replace the ball $\B_{\epsilon}(x)$ with some parameterized set $\pi_{\epsilon}(x)$. 
For a set of labelled training data $\{ (x_i,y_i)\}_{i=1}^{n}$, robust
training can be formulated as a saddle-point problem:
\begin{equation}
\underset{\theta}{\mathrm{min}}\; \underset{i}{\mathrm{max}} \;
    \L_{N_{\theta}}(x_i, y_i)
\label{eq:robust-training}
\end{equation}

Finding the worst-case adversarial loss for a given example is
computationally expensive in general. 
In practice, most approaches approximate the worst-case adversarial 
loss in one way or 
another~\cite{provable-defenses,wong2018scaling,pmlr-v80-mirman18b}.
In recent years, robust training has progressed from 
single layer, dense networks to moderate --- though 
not yet state-of-the-art --- sized \acp{cnn}.
This has brought these techniques within the realm of various
\acp{dnn} used within the reference \ac{av} stack being built by \five.
In our bid to understand the practicalities of robust training, we 
found that existing tools are inconvenient to use or apply to existing models: they typically only support a small subset of model 
elements and require users to re-specify the models in a specialized
language, which can mean extensive rewrites to the training code.

To tackle these problems, we introduce a framework in this paper, called
\emph{P}r\emph{a}ctical \emph{Ro}bust \emph{T}raining
(\framework)\footnote[1]{The framework is  
available at \url{https://github.com/fiveai/parot}},
developed on the popular TensorFlow platform~\cite{abadi2016tensorflow}. 
Our framework allows robust training ---  using differentiable abstract 
interpretation~\cite{pmlr-v80-mirman18b} --- to be performed on arbitrary 
\acp{dnn} without any rewrites of the model. 
In \framework, one can start a robust model training for a popular
convolutional neural network with a minimal amount of code, as we 
demonstrate in Listing~\ref{lst:training_code}. We have, for example,
used \framework to robustly train the traffic light detection network
seen in Figure~\ref{fig:traffic-light}.

\paragraph{Contributions}
The main contribution of this paper is a practical framework, \framework,
built in the Tensorflow platform~\cite{abadi2016tensorflow}. 
In particular,
\begin{compactitem}
\item Our tool can automatically apply abstract interpretation on an 
existing model definition. 
Thus, it can be used to verify robustness on existing \acp{dnn} without 
having to change the model code, allowing for seamless adoption with
existing codebases.
\item Our framework implements a broad set of robustness properties that
go beyond the usual $\epsilon$-ball, and provides a clean interface for
specifying custom properties.
\item We improve upon the abstract interpretation techniques
used by Mirman \etal~\cite{pmlr-v80-mirman18b}. In particular, we refine several
abstract transformers for activation functions.
\end{compactitem}

\paragraph{Structure of paper}
In Section~\ref{s:background} we introduce the requisite
background in robust training with abstract interpretation.
In Section~\ref{s:system} we describe the architecture and
functionality of the \framework framework and evaluate its performance
in Section~\ref{s:experiments}.
In Section~\ref{s:related} we place our work more broadly 
in the field of formal verification of \acp{dnn}. Finally, in
Section~\ref{s:conclusion}, we conclude and present future directions
for this framework and paradigm.

\section{Background}
\label{s:background}

We build on the robust training approach of DiffAI, introduced by Mirman \emph{et al.}~\cite{pmlr-v80-mirman18b}, where the inner maximization of 
Equation~\ref{eq:robust-training} is approximated using abstract
interpretation. 
In this section, we sketch the mathematical prerequisites to our
framework.

\subsection{Abstract Interpretation}
Abstract interpretation is a general theory for approximating
infinite sets of behaviours with a finite
representation~\cite{cousot1977abstract,cousot1992abstract}.
In the present study, this corresponds to convex approximations 
of a non-convex adversarial polytope.

\begin{figure}[ht]
\centering
\includegraphics[width=0.8\columnwidth]{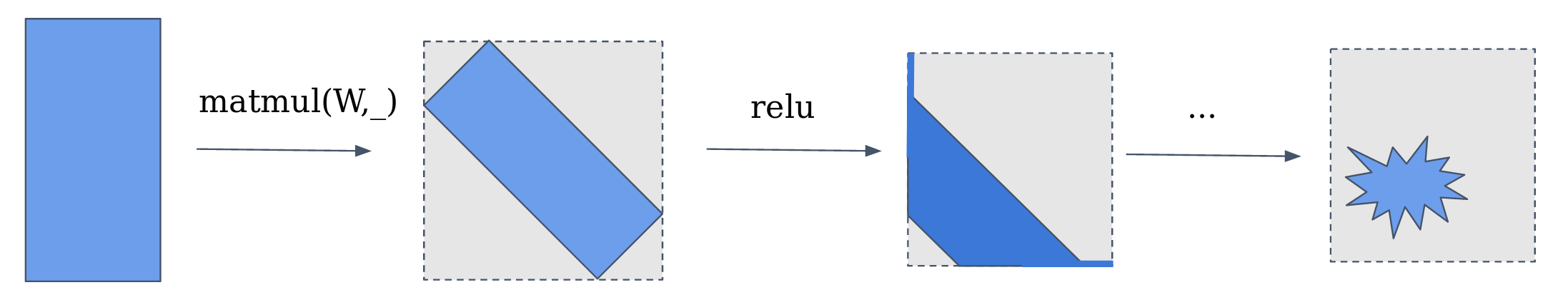}
\caption{An illustration of abstract domains. The dotted grey box corresponds to a domain object, and the blue shape is the true set that the domain object seeks to approximate.
}
\label{abstract-domain}
\end{figure}

The two basic constructs in abstract interpretation are the 
\emph{abstract domain} and the \emph{abstract transformer}.
Intuitively, an abstract domain gives a finite (approximate)
representation of the (potentially infinite) concrete space, 
while an abstract transformer provides an over-approximation
of the behaviour of a function.
Formally, an abstract domain is a set $\D$ (the \emph{domain}) 
and a pair of maps $\alpha : \P{\real^p} \to \D$ and
$\gamma : \D \to \P{\real^p}$, called the \emph{abstraction} 
and \emph{concretization} maps, respectively. $\P{X}$ is the 
powerset of $X$.
The abstraction function is defined such that
$U \subseteq \gamma (\alpha(U))$ for all $U \subseteq \real^p$.

Additionally, an abstract domain is equipped with a mapping 
from a fixed set of primitive functions $\F$ to \emph{abstract 
transformers} in $\D$ such that each 
$f: \real^p \to \real^q$ in $\F$ is mapped to a function 
$\D(f) : \D \to \D'$.
For each element in the concrete space, $z$, transformers must obey 
the following \emph{soundness relation}:
\begin{equation}\label{sound} f[\gamma(z)] \subseteq \gamma(\D(f)(z))
\end{equation}
This ensures that transformers produce new abstract elements
whose concretization overapproximates the image of the function.
Since transformers compose, we may transform any composite 
function $f = f_1 \o f_2 \o \cdots \o f_n : \real^p \to \real^q$ 
where $f_i \in \F$.
Figure~\ref{abstract-domain} illustrates graphically the abstract 
domains and transformers for a single layer of a \ac{dnn}.
We can construct a composite transformer
$\D(N)$ that represents that network, and write the sound 
approximation for an $\epsilon$-ball around a point $x$
as:
\begin{equation}
\gamma(\D(N)[\alpha(\B_{\epsilon}(x))]).
\end{equation}
\subsection{Abstract Domains for \acp{dnn}}
\label{sec:abstraction-domains}
We consider three abstract domain types:
\bx, \zono and \hz:
\begin{compactitem}
    \item \bx, represented by $i = \langle\c, \b \rangle$. A \bx 
    domain is a $p$-dimensional axis-aligned box, parameterized by 
    its center $\c \in \real^p$ and a positive vector 
    $\b \in \real^p_{>0}$
    containing the half-widths of the box. Figure \ref{abstract-domain}
    illustrates the concept of the \bx domain.
    \item \zono, represented by $z = \langle\c,\E \rangle$. For dimension $p$,
    a \zono is parameterized by a center point $\c \in \real^p$ as well as a
    matrix $\E \in \real^{p \times e}$ for some fixed dimension $e$. 
    The set $z \subseteq \real^p$  is the $\E$ image of an
    $e$-dimensional hypercube, centerd at $\c$. The concretization
    is given by:
    \begin{equation}
    \gamma(z) := \{\c + \E \; \v \; : \;  \abs{\v_i} \leq 1, \;
    i \in \{1,\ldots,e\}\}
    \end{equation} 
    The key feature of a \zono domain is that transformers
    exist for affine functions --- such as the matrix 
    multiplications associated with transition functions of
    \acp{dnn} --- that do not increase the approximation error. 
    \item \hz, represented by $h = \langle \c,\b,\E \rangle$. One problem
    with the \zono domain is that computation can be expensive
    compared to a \bx domain. The \hz solves this
    problem with the inclusion of an extra positive vector $
    \b \in \real^p_{>0}$, with a concretization:
    \begin{multline}
    \gamma(h) := \{\c + \E \; \v + \mathrm{diag}(\b) \; \w  \;|\; 
    \abs{\v_i} \leq 1, \; \abs{\w_j} \leq 1, \\
    i, \in \{1,\ldots,e\}, \; j \in \{1, \ldots, p\}\}.
    \end{multline}
\end{compactitem}

 Note that these definitions mean that \bx and \zono are both subsets of \hz.
 In the \hz domain, it is possible to convert $\b$ values to $\E$ values and 
 vice-versa through  \emph{correlation} and
 \emph{decorrelation}, as noted in~\cite{deep-residual}.

\subsection{Hybrid Zonotope Transformers for \acp{dnn}}
It is straightforward to show that exact transformers can be 
constructed for matrix multiplication~\cite{pmlr-v80-mirman18b}.
In contrast, accurate modeling of piecewise linear activation 
functions, such as
$
\mathrm{relu}(x) := \max(x,0)
$, necessarily introduce an approximation.
Here we generalize the work in~\cite{singh2018fast} to find optimal 
hybrid zonotopes for a given activation function. 
Since activations are one-dimensional (1D) and act on each dimension
separately, we may consider just the problem in 1D. 
For a given function $f : \mathbb{R} \to \mathbb{R}$ and 
input bounds $\ux$, $\ox$, the challenge is  to find a parallelogram containing the graph of $f$ 
restricted to $[\ux,\ox]$ that has minimal area, as shown in
Figure~\ref{constructzono} below.

In the first instance, we consider an activation function $f$ which is 
convex or concave.
If $\ox = \ux$, we can treat the transformer as acting on a point.
Otherwise, we compute the slope of the parallelogram:   
$$\mu := \frac{f(\overline{x}) - f(\underline{x})}{\overline{x} - \underline{x}}$$
\noindent We provide an \emph{extremum function} $x_f(\mu)$ for 
the given $f$. Assuming a convex function:
\begin{equation}
x_f(\mu) = \argmin{x \in \mathbb{R}} (f(x) - \mu x)
\end{equation}
If $f$ is concave, replace $\amin$ with $\amax$. 
Since $f$ is convex/concave, this $x_f(\mu)$ will always be in the 
interval $[\ux,\ox]$ or otherwise $f(x) - \mu x$ is zero everywhere 
in $[\ux,\ox]$.
For many of the activation functions we care about, it is simple to find these extremum 
functions.
For example, $x_{\mathrm{relu}}(\mu) = 0$ and $x_{\exp{}}(\mu) = \ln{\mu}$.
Then, one can compute:
\begin{equation}
e := x_f(\mu) \cdot \mu - \frac{f(\ux) \cdot \ox + \ux \cdot f(\ox)}{\ox - \ux}
\end{equation}
\noindent which may be interpreted as the height of the resulting zonotope
parallelogram. From this we may compute the center of the parallelogram in 
the $y$ direction:

\begin{equation}
c_y := \frac{1}{2} (f(\ox) + f(\ux) - e)
\end{equation}

\noindent Finally we compute the new 1D hybrid zonotope:

\begin{equation}
D(f)\langle c_x, b_x, \E_x\rangle = \langle c_y, \; \mu b_x + \frac{e}{2}, \; \mu \E \rangle
\end{equation}

To extend this approach to nonconvex functions, such as $\mathrm{sigmoid}$,
we instead need to find a pair of extrema $\underline{x_f}(\mu)$,
$\overline{x_f}(\mu)$ which may in general  depend on the interval 
bounds $[\ux,\ox]$. 
In the case of $\mathrm{sigmoid}$, one can show that these are minus the 
natural logarithm of the solutions $Y_\pm$ to the quadratic equation 
$\mu + (2\mu - 1)Y + \mu Y^2 = 0$.
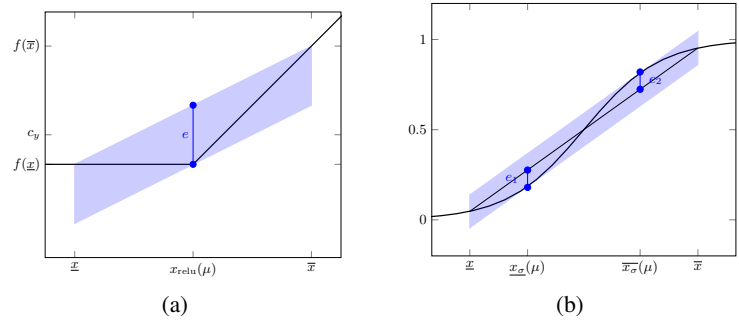
\begin{figure*}[t!] 
    \centering
    \subfloat[\label{fig:relu-zonotope}]{
        \centering
        \resizebox{0.37\columnwidth}{!}{
        \begin{tikzpicture}
          \begin{axis}[ xmin = -2.5, xmax = 2.5, ymin = -1.5, ymax = 2.5, xtick = {-2,0,2},
                       xticklabels = {$\ux$, $x_{\mathrm{relu}}(\mu)$, $\ox$},
                       ytick       = {0, 0.5,2},
                       yticklabels = {$f(\ux)$,$c_y$, $f(\ox)$},
                       axis equal,
                       set layers,
                      ]
            \addplot+[mark=none, thick, black,domain=-5:0] {0};
            \addplot+[mark=none, thick, black,domain=0:5] {x};
            \addplot+[mark=none, fill, opacity=0.2, blue] coordinates {(-2,0) (2,2) (2,1) (-2,-1)} --cycle;
            \addplot+[mark=*, blue] coordinates {(0,0) (0,1.0)} [xshift=-5pt] node[pos=0.5] {$e$};
            \addplot+[mark=arrow] coordinates {(0,0.5)};    
          \end{axis}
        \end{tikzpicture}
        }
    }%
    \hspace{0.5cm}
    \subfloat[\label{fig:sigmoid-zonotope}]{
        \centering
        \resizebox{0.37\columnwidth}{!}{
        \begin{tikzpicture}
        \begin{axis}[
               xmin    = -4,
               xmax    = 4,
               ymin    = -0.2,
               ymax    = 1.2,
               xtick       = {-3, -1.48,  1.48, 3},
               xticklabels = {$\ux$, $\underline{x_\sigma}(\mu)$, $\overline{x_\sigma}(\mu)$ , $\ox$},
               set layers,
              ]
            \addplot+[mark=none, thick, black,domain=-5:5] { 1 / (1 + exp(-x))};
            \addplot+[fill, opacity=0.2, blue, mark=none] coordinates {(-3,-0.0439) (-3,0.138) (3,1.0438) (3,0.8612)} --cycle;
            \addplot[mark=none, black] coordinates {(-3,0.047) (3,0.953)} [mark=none];
            \addplot[mark=*, blue] coordinates {(-1.48,0.18) (-1.48,0.276)} [xshift=-10pt] node[pos=0.5] {$e_1$};
            \addplot[mark=*, blue] coordinates {(1.48,0.82) (1.48,0.724)}   [xshift=10pt]  node[pos=0.5] {$e_2$};      
        \end{axis}
        \end{tikzpicture}
        }
    }
    \caption{Constructing zonotope transformers for $\mathrm{relu}$ and 
    $\mathrm{sigmoid}$ activation functions.}
    \label{constructzono}
\end{figure*}
Figure~\ref{constructzono} shows zonotope transformers for $\mathrm{relu}$ and 
    $\mathrm{sigmoid}$ activation functions.

\subsection{Robust Training}

To train with an abstract domain on a model $N$, from each training
datum $(\x,\y)$ we compute a prediction value $N(\x)$ and a 
transformed domain object $\D(N)(\B_\epsilon(\x))$ of the 
domain representation of an $\ell_\infty$-ball
$\B_\epsilon(\x)$ around the input $\x$ for some fixed
perturbation radius $\epsilon$.
An axis-aligned bounding box is drawn around the resulting
output domain object, and the vertex $v$ furthest away 
from the true target $\y$ is chosen.
We construct a combined loss $\L_{\mathrm{comb}}$ with the standard loss, the adversarial loss, a mixing factor $\lambda \in \real_{\geq 0}$, and a regularization term $\xi(N)$:
\begin{equation}
\label{eq:comb-loss}
    \L_{\mathrm{comb}}(\x,\y) \\
        := \L(N(\x),\y) 
            + \lambda \L(\argmax{\v \in \D(N)(\x)} \norm{2}{\v - \y}, \y) + \xi(N)
\end{equation}

\section{PaRoT System Description}\label{s:system}
\begin{figure}[t]
\centering
\includegraphics[width=0.65\columnwidth]{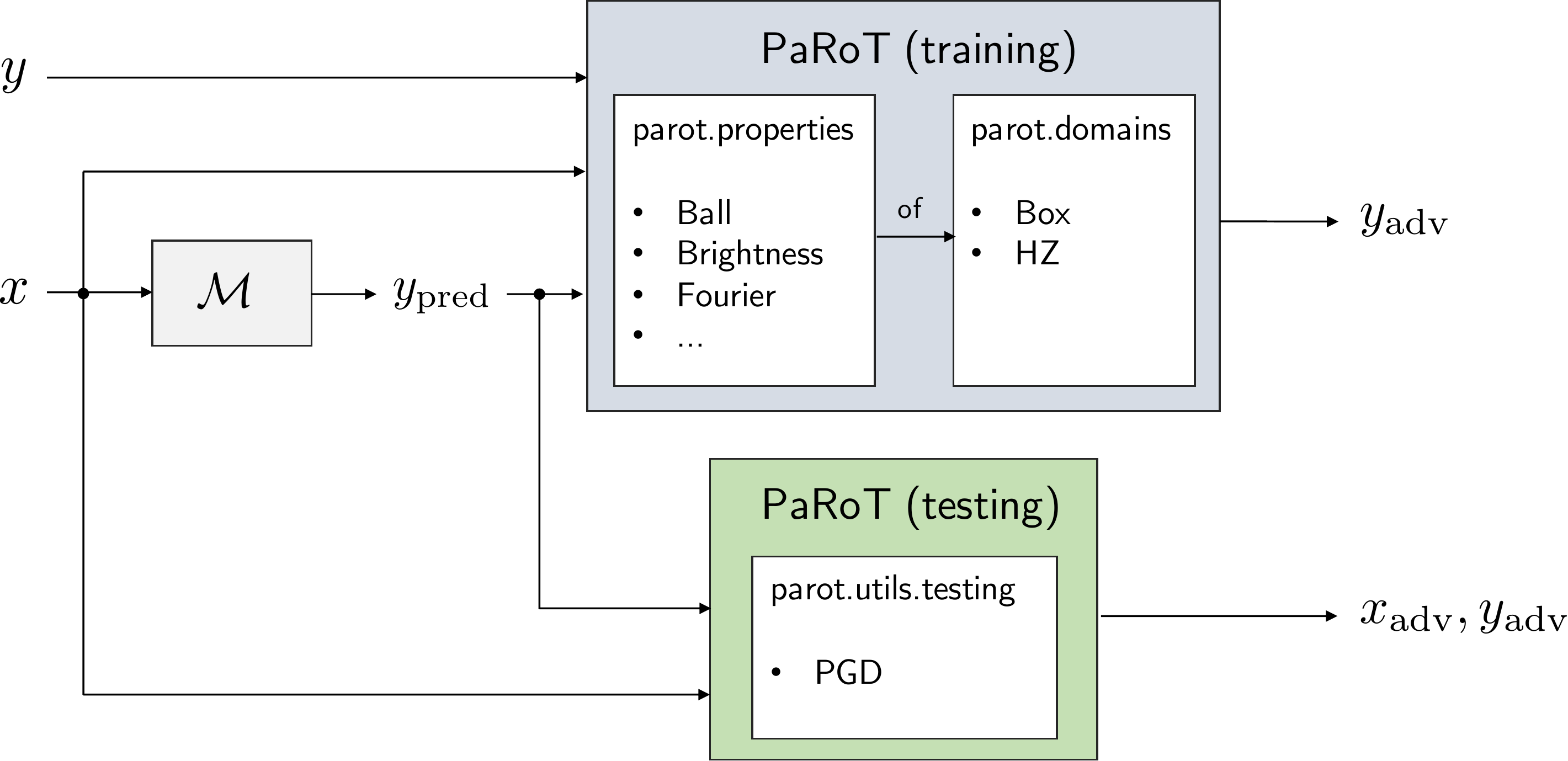}
\caption{\framework overview. Robust training is enabled by a property and an abstraction domain which can be chosen from those supported by \framework or extended with a custom domain. Given an input and a model prediction, \framework creates a domain object for the input based on the specified property, and it automatically transforms the operations associated with the model (see Section~\ref{sec:auto_transform}). At testing time, \framework provides auxiliary utilities.}
\label{fig:parot_overview}
\end{figure}

In this section, we detail how \framework can be used for robust training and testing. The main overview of the system is presented in Figure~\ref{fig:parot_overview}. The training aspects of the framework can be divided into \textit{domains} (in the module \texttt{parot.domains}), which correspond to the ones identified in Section~\ref{sec:abstraction-domains}\footnote{With the exception of the \zono domain, which is not implemented in \framework.}, and \textit{properties} (in the module \texttt{parot.properties}) corresponding to the types of adversaries we are trying to robustify against. Section~\ref{sec:properties} presents the built-in properties available in \framework.
As our system uses the TensorFlow platform, we first introduce
some terminology.

TensorFlow~\cite{abadi2016tensorflow} is a deep learning platform that enables the user to build a \emph{computation graph} representing their neural network model 
and training scheme.
This computation graph is a directed, acyclic graph 
whose nodes are \emph{{\Tensor}s} --- a generalization of matrices
to potentially higher dimensions --- and whose edges are called \emph{{\Op}s} and consist 
of a list of input and output {\Tensor}s. 
An output \Tensor can be the 
input \Tensor for arbitrarily many {\Op}s.
To illustrate, the left-hand side of Figure~\ref{transformer_flow_chart} shows
the computation graph constructed for a single dense layer of a neural network.
The {\Op}s $\mathrm{MatMul}$ (matrix multiplication), $\mathrm{BiasAdd}$ (adding a bias to a value), and $\mathrm{ReLU}$ (rectified linear unit operation)
form those required to represent this example layer. 
Once a computation graph has been created, TensorFlow compiles it, allowing
\framework to use this graph to automatically derive abstract transformers for a
given model, as described in Section~\ref{sec:auto_transform}.
This enables a user to use an existing model and
immediately start robust training without needing code rewrites. It should be noted that
the models supported by \framework must use only the operations supported by the framework
in the selected domain. A list of the operations is available in Appendix~\ref{app:tf-ops}.
\subsection{Automatic Transformer Generation}
\label{sec:auto_transform}

In order to transform a computation graph $G$ from a given input \Tensor $\x$ to 
an output tensor $\z$, we find the subgraph $S_{\x,\z}$ of $G$ whose vertices 
are the $\y$s such that there exist paths $\x \leadsto \y$ and $\y \leadsto \z$.
This can be easily extended to multiple inputs and outputs.
This subgraph $S_{\x,\z}$ is found through a graph traversal 
algorithm backtracking from $\z$, which also produces a pair of adjacency maps 
$C$ and $M$. $C$ maps a tensor to a set of {\Op}s which consume it, while $M$ maps an \Op $f$ to the indices of the output tensors of the \Op  in $G$.
Once $S_{\x,\z}$ is constructed, the transformation process can begin.
The output of the process is a dictionary  $\mathcal{T}$ which maps $p$-dimensional tensors to domain objects
$D$ (or the constant $\mathtt{None}$). $\mathcal{T}$ is constructed by iteratively exploring
$S_{\x,\z}$ starting at $\x$. 
The complete transformation algorithm
is given in Algorithm~\ref{alg:graph_transform}.

\begin{algorithm}[h]
\SetAlgoLined
\KwData{A subgraph $S_{\x,\z}$, initial domain object $D \in \D$}
\KwResult{A transformed domain object $\Z \in \D$}
front $\leftarrow [\x]$\;
$\mathcal{T} \leftarrow \{\x \mapsto D\}$\;
\While{front $\neq \emptyset$}{
$\s \leftarrow$ pop front\;

\For{$f \leftarrow $ consumers of $\s$ in $S_{\x,\z}$} {
$\s_1, \cdots, \s_n \leftarrow$ the inputs of $f$ in $G$\;
\If{$\exists \; i$: $\s_i \in S_{\x,\z} \wedge \s_i \notin \mathcal{T}$} {
continue\tcp*{wait for other inputs to be transformed}
}
\For{$1 \leq i \leq n$}{
\eIf{$\s_i \notin \mathcal{T} \vee \mathcal{T}(\s_i) = \mathtt{None}$} {
$\S_i \leftarrow \s_i$\;
}{
$\S_i \leftarrow \mathcal{T}(\s_i)$\;
}
}
$\T_1,\cdots, \T_m \leftarrow \D(f)(\S_1, \cdots, \S_n)$\;
\For{$1 \leq j \leq m$} {
$\j \leftarrow j$th output of $f$\;
$\mathcal{T}(\j) \leftarrow T_j$\;
push $j$ to front\;
}
}
}
\Return{$\mathcal{T}(\z)$;}
\caption{Automatic graph transformation algorithm}
\label{alg:graph_transform}
\end{algorithm}

When transforming {\Op}s, various challenges arise. For example, a transformer $\D(f)$ can  accept inputs that are not domain objects but
instead just tensors. 
This occurs, for example, when a constant tensor needs to be added to a domain
object.
The acyclic graph structure makes this transformation non-trivial. The first 
issue arises when an operation consumes two or more domain objects. This happens in reticulated model architectures \eg \texttt{SkipNet} from \cite{wang2018skipnet}. 

To illustrate the challenges of transforming ops, take two
{\Tensor}s $\x$, $\y$, consider the transformed computation graph
for their addition $\x + \y$ where both 
$\x$ and $\y$ have abstract domains to be transformed. 
To transform $+$ for the \bx domain, this entails merely adding the $\c$s 
and $\b$s of $\x$ and $\y$.
However, for \hz, the manner with which the merging should take place
depends on how the $\E$ matrices were constructed. 
If $\x$ and $\y$ are both derived from the same starting zonotope, then their 
$\E$ matrices will both be referencing the same parameterization. 
In this case the $\E$ matrices for $\x$ and $\y$ can be added.
However, if they originate from different starting zonotopes, then their $e$ dimensions may not match up, and in
this case they need to be \emph{concatenated} along the $e$ dimension:
\begin{equation}
    \langle \c_\x, \b_\x, \E_\x \rangle 
    + \langle \c_\y, \b_\y, \E_\y \rangle
    := \langle \c_\x + \c_\y, \b_\x + \b_\y, [\E_\x, \E_\y] \rangle
\end{equation}
Similar considerations must be made for, e.g., the $\mathtt{Concat}$ \Op which 
concatenates two tensors along a given dimension. 

\begin{figure}[H]
\centering
\includegraphics[width=0.65\columnwidth]{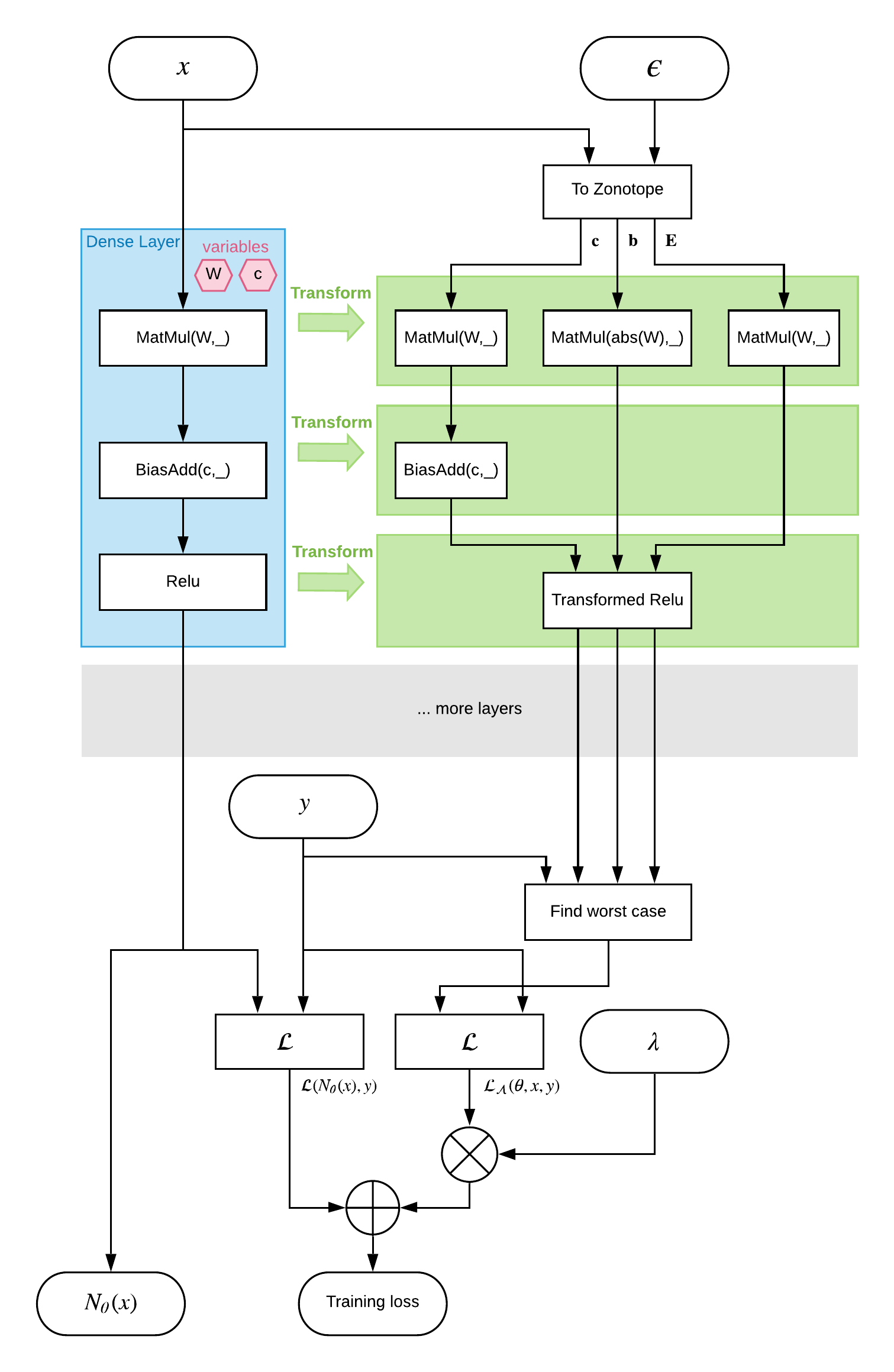}
\caption{An example computation graph for robust training showing the original (blue) and generated (green) computation graph on a dense layer. }
\label{transformer_flow_chart}
\end{figure}

Another complication in extending transformers to computation graphs arises with {\Op}s which do not output a transformed domain object. 
The most prominent example of this is the \texttt{Shape} \Op which returns the
dimensions of a tensor. 
We support these operations by allowing the domain implementer to return
\texttt{None} instead of a domain object, flagging that the transformer algorithm
should use the un-transformed output.

With these two considerations in mind, we have a procedure for transforming
arbitrary TensorFlow graphs composed from a set of atomic transformers.

Figure~\ref{transformer_flow_chart} illustrates 
computing the transformed graph of the nodes on the left-hand side which represent the {\Op}s of a 
dense layer.
Each green group on the right-hand side is the generated transformer computation graph
of the corresponding $f$ in the domain \hz, 
\ie, the result of calling $\D(f)$ for the \Op. 
Note that the variables from the original layer are shared with the transformed {\Op}s.

\lstset{style=lst_style}
\lstset{language=Python}
\begin{lstlisting}[float,escapeinside={(*}{*)},
caption={Implementing custom properties with \framework.}, 
                        label={lst:custom_prop}]
from parot.domains import Box
from parot.properties import Property

class NewProperty(Property):
    # define the supported domains of this property
    SUPPORTED_DOMAINS = [Box]
    
    def __init__(self, param)
        # replace the initializer to accept desired parameters
        pass
    
    def generate_property(self, domain, input_tensor):
        # implement the property here; safely assume domain is one
        # of the types in NewProperty.SUPPORTED_DOMAINS
        pass

# use the property on a tensor
x_box = NewProperty(param_instance).of(Box, x)
\end{lstlisting}

\subsection{Robustness Properties}
\label{sec:properties}
 In this section, we describe several built-in robustness properties that
can be trained with in \framework,
and an interface for specifying custom properties.

\paragraph{Built-in Properties}
Let $\mathbb{1}_s$ denote a tensor with shape $s$ with all elements being ones. All the following supported properties are centered on a training input $x$ with shape $s$.
\begin{compactitem}
    \item \texttt{BallDemoted}: the $\ell_{\infty}$-norm ball adversarial attack represented as an axis-aligned \textsc{Box} where $\b = \epsilon \cdot \mathbb{1}_s$.
    
    \item \texttt{BallPromoted}: another $\ell_{\infty}$-norm ball adversarial attack represented in the $\E$ matrix of the \textsc{HybridZonotope} as $\E = \epsilon \cdot \mathrm{diag}(\mathbb{1}_{s})$
    
    \item \texttt{Brightness}: a simple property with a single column in $\E$ where all 
        pixels may have a constant added to them. That is, $\E = \epsilon \cdot \mathbb{1}_{\dots s,1}$.
    
    \item \texttt{UniformChannel}: similar to \texttt{Brightness} except that each
        channel of the image is allowed to vary independently.
    
    \item \texttt{Fourier}: for a 2D image $x$, each column of $\E$ is a plane wave.
        That is, each column of $\E$ is an image $I: H\times W \to \real$:
        $$I(i,j) = \epsilon \cdot \kappa\left(i\frac{2 \pi n}{H} + j\frac{2 \pi m}{W}\right),$$
        for $\kappa \in \{\sin, \cos\}$, $n \in \{-N,...,N\} \subset \mathbb{Z}$ and $m \in \{-M,...,M\} \subset \mathbb{Z}$.
        Our motivation to investigate this property is to study the robustness to perturbations that we might observe in real data collected 
        in the field.
        For example, in the case of detecting traffic lights, we can investigate 
        whether it is possible to attack the network using only low frequencies (to model 
        markings or distortions on a physical traffic light).
        An example of an adversarial example obtained through the $\mathtt{Fourier}$ on \texttt{MNIST}~\cite{mnist} is 
        shown in Figure~\ref{fig:fourier-example}.
\end{compactitem}

\paragraph{Custom Properties}

Defining a custom property in \framework is as simple as implementing a child class of \texttt{Property}, as presented in Listing~\ref{lst:custom_prop}.

\subsection{Robust Training using \framework} 
Integrating our framework in a codebase can easily be done with minimal changes to the
existing code, as exemplified in Listing~\ref{lst:training_code}. 
Given a training dataset with inputs \texttt{x} and groundtruth outputs \texttt{y}
in \Tensor form, as well as the predictions of the model for the inputs,
\texttt{y\_pred}, we create a domain object using a \textsc{Box} abstraction around the inputs and transform the resulting computation graph. 
Then, a combined loss function can be created and passed to the 
desired optimizer for robust training.

\lstset{style=lst_style}
\lstset{language=Python}
\begin{lstlisting}[float,escapeinside={(*}{*)},
caption={Given a model graph, return the 
                    training operation that optimizes a weighted version of the loss function for adversarial
                    training.}, 
                        label={lst:training_code}]
from parot.domains import Box
from parot.properties import Ball

# get the epsilon ball around the input in a given domain
x_box = Ball((*$\epsilon$*)).of(Box, x)

# transform the graph and obtain the output box
y_box = x_box.transform(outputs=[y_pred], input=x)
y_adversary = y_box.get_adversary(y)

# create the combined loss function
regular_loss = loss_function(y, y_pred)
adversary_loss = loss_function(y, y_adversary)
combined_loss = regular_loss + (* $\lambda$ *) * adversary_loss

# obtain the training operation
train_op = optimizer.minimize(combined_loss)
\end{lstlisting}

\section{Experiments}\label{s:experiments}
We evaluate \framework quantitatively to demonstrate
performance, and qualitatively to validate its ease of use. 
We first show that our performance is comparable to the results obtained by DiffAI~\cite{pmlr-v80-mirman18b}. 
We then exemplify the ease of use on pre-trained models and finish with qualitative examples demonstrating a \framework robustness property.
Throughout these experiments, we use the terms `standard', `regular' and `baseline'
interchangeably to describe a training process that solely uses a sparse cross-entropy loss. 

In quantitative experiments, we make use of three metrics to measure performance: 
\begin{compactitem}
    \item \textbf{Test Error}: percentage of misclassified examples in the testing set; the complement of classification accuracy.
    \item Test error under a \textbf{PGD} attack: a test based on the state-of-the-art Projected Gradient Descent attack first presented in~\cite{MadryResistant}
    and used in~\cite{pmlr-v80-mirman18b}. PGD finds an adversarial example by  following the gradient of the loss function inside an $\epsilon$-ball around the actual test example on the input side.  Thus, the reported values correspond to a lower bound on the percentage of the misclassified examples in the testing set that are susceptible to an attack of this type.
    \item Test error under a \hz attack \textbf{Verify}: similar to the \texttt{hSwitch} upper bound metric
    in~\cite{pmlr-v80-mirman18b}, this metric uses the adversarial example discovered by the \hz on the output side, as in \eqref{eq:comb-loss}.
    Thus, the reported values correspond to an upper bound of the 
    percentage of verifiably-susceptible examples in the testing set under this attack.
\end{compactitem}

\subsection{DiffAI Comparison}
To validate the results of our framework, we ran robust training experiments similar to those in~\cite{pmlr-v80-mirman18b} of the \bx and \hz domains for
\texttt{MNIST}~\cite{mnist} and \texttt{CIFAR10}~\cite{cifar10}.
The architecture of the networks used is as defined in~\cite{pmlr-v80-mirman18b} and is also presented in Appendix~\ref{app:networks}. 
We similarly augment  the loss with an adversarial term with weight $\lambda=0.1$ and an \textit{L2}
regularization constant of $0.01$. The learning rate and $\epsilon$ used are $10^{-3}$ and $0.1$ 
for \texttt{MNIST}, and $10^{-4}$ and $0.007$ for \texttt{CIFAR10}, 
respectively. 
We run all experiments for $100$ epochs using the Adam optimizer \cite{adam}. The 
results are in Table~\ref{tab:diffai-comp}.

As the  table shows, our framework achieves comparable 
results to those obtained in~\cite{pmlr-v80-mirman18b}. 
In all cases, introducing an adversarial training method leads to a minor
drop in accuracy --- an expected outcome when optimizing for a 
combined loss function with a finite 
capacity~\cite{tsipras2018robustness,friendslikethese}. 
In terms of \textsc{Box} training, we observe, as expected, a slight increase in PGD and a strong increase in the number of verifiably-safe examples. 
For a \hz training when compared to the baseline, we notice that the number of 
examples susceptible to a PGD attack grows slightly while, in general, the number 
of verifiable cases improves significantly. 
Overall, these results are similar and in many cases improve upon the ones in
\cite{pmlr-v80-mirman18b} with minor exceptions that can be justified by 
implementation differences and stochasticity in weight initialization. 

\begin{table}
\centering
\bgroup
\def\arraystretch{1.3}
\begin{tabularx}{\textwidth}{lll|*3{>{\centering\arraybackslash}X}@{}}
\textbf{Dataset} \hspace{1.5em} & \textbf{Model} \hspace{2.5em} & \textbf{Train Method}\hspace{1.5em} & \textbf{Test Error} \%               & \textbf{PGD} \%          & \textbf{Verify} \% \\ \hline
\multirow{15}{*}{\texttt{MNIST}}  &   \multirow{3}{*}{FFNN}  &   Baseline    &   1.8 &   3.2 &   100.0 \\ 
  &     &   \textsc{Box}    &   3.2 &   4.2 &   30.6 \\
  &     &   \textsc{HybridZonotope}    &  3.2 &   4.0 &   30.2 \\ \cline{2-6}
  &   \multirow{3}{*}{ConvSmall}  &   Baseline    &   1.4 &   2.4 &   100.0 \\ 
  &     &   \textsc{Box}    &  2.0 &   2.4 &   12.8 \\
  &     &   \textsc{HybridZonotope}    &   1.8 &  2.4 &   91.8 \\ \cline{2-6}
  &   \multirow{3}{*}{ConvMed}  &   Baseline    &   1.8 &   2.2 &   100.0 \\ 
  &     &   \textsc{Box}    &   1.8 &   2.2 &   13.6 \\
  &     &   \textsc{HybridZonotope}    &   2.4 &   2.6 &   88.6 \\ \cline{2-6}
  &   \multirow{3}{*}{ConvBig}  &   Baseline    &   0.6 &   1.2 &   100.0 \\ 
  &     &   \textsc{Box}    &   1.2 &   1.4 &   14.0 \\
  &     &   \textsc{HybridZonotope}    &   1.8 &   2.0 &   74.2 \\ \cline{2-6}
  &   \multirow{3}{*}{ConvSuper}  &   Baseline    &   0.6 &   1.0 &   100.0 \\ 
  &     &   \textsc{Box}    &   1.0 &   1.2 &   12.2 \\
  &     &   \textsc{HybridZonotope}    &   1.0 &   1.6 &   72.4 \\ \cline{2-6}
  &   \multirow{3}{*}{Skip}  &   Baseline    &   0.6 &   0.8 &   100.0 \\ 
  &     &   \textsc{Box}    &   1.0 &   1.8 &  11.0 \\
  &     &   \textsc{HybridZonotope}    &   0.8 &   1.6 &   10.0 \\ \hline \hline
\multirow{15}{*}{\texttt{CIFAR10}}  &   \multirow{3}{*}{FFNN}  &   Baseline    &   45.8 &   45.8 &   100.0 \\ 
  &     &   \textsc{Box}    &   50.4 &   50.4 &   76.2\\
  &     &   \textsc{HybridZonotope}    &   48.8 &  48.8 &   75.8 \\ \cline{2-6}
  &   \multirow{3}{*}{ConvSmall}  &   Baseline    &   33.3 &  33.4 &   100.0 \\ 
  &     &   \textsc{Box}    &   36.2 &   36.2 &   72.0 \\
  &     &   \textsc{HybridZonotope}    &   38.6 &  38.6 &   96.2 \\ \cline{2-6}
  &   \multirow{3}{*}{ConvMed}  &   Baseline    &  34.6 &   34.6 &   100.00 \\ 
  &     &   \textsc{Box}    &   35.8 &   35.8 &   69.6 \\
  &     &   \textsc{HybridZonotope}    &   34.4 &   34.6 &   96.4 \\ \cline{2-6}
  &   \multirow{3}{*}{ConvBig}  &   Baseline    &   35.4 &   35.6 &   100.0 \\ 
  &     &   \textsc{Box}    &   36.0 &   36.0 &   71.2 \\
  &     &   \textsc{HybridZonotope}    &   38.0 &  38.0 &   99.4 \\ \cline{2-6}
  &   \multirow{3}{*}{ConvSuper}  &   Baseline    &   34.4 &   35.2 &   100.0 \\ 
  &     &   \textsc{Box}    &   33.6 &   34.2 &   100.0 \\
  &     &   \textsc{HybridZonotope}    &   35.3 &   35.4 &   98.6 \\ \cline{2-6}
  &   \multirow{3}{*}{Skip}  &   Baseline    &   34.0 &   34.6 &   100.0 \\ 
  &     &   \textsc{Box}    &   40.0 &   39.8 &   73.2 \\
  &     &   \textsc{HybridZonotope}    &   39.4 &  39.6 &  74.0 \\ \hline
\end{tabularx}
\egroup
\vspace{0.5em}
\caption{\textit{Quantitative Comparison}: results of running our framework on the same datasets, architectures and parameters as in~\cite{pmlr-v80-mirman18b}. In the experiments run, we used $\epsilon=0.1$ for \texttt{MNIST} and $0.007$ for \texttt{CIFAR10}.}
\label{tab:diffai-comp}
\end{table}

\subsection{Re-training Models}
In this experiment, we showcase the ease of use of \framework using a pre-trained 
network. 
We train a network with two convolutional layers and two dense layers, following the
architecture of \texttt{ConvSmall} (see~\cite{pmlr-v80-mirman18b}), on the 
\texttt{MNIST} dataset using a standard loss for $200$ epochs (learning rate of $10^{-3}$)
and save it to a TensorFlow checkpoint file. 
We proceed to load this checkpoint's graph, and, using \framework's \textsc{Box}
abstract domain, add an adversarial term to the loss function, which we then use to
further train the loaded model for $100$ epochs. The results of the process are 
presented in Table~\ref{tab:re-train}. 
Re-training achieves similar accuracy, while improving significantly the PGD and
verification metrics. 
It should be noted that at no point in the re-training process did we have to re-define
the model or state the required operations, one of the main advantages of our
framework.

\begin{table}[t]
\centering
\bgroup
\def\arraystretch{1.3}
\begin{tabularx}{0.75\textwidth}{l|*3{>{\centering\arraybackslash}X}@{}}
\textbf{Model} \hspace{6em} & \textbf{Test Error} \%    & \textbf{PGD} \%   & \textbf{Verify} \% \\ \hline
Original                    & 1.70       & 2.30       & 100.00 \\ 
Re-trained (\textsc{Box})   & 2.88       & 1.47       & 14.80  \\ 
\end{tabularx}
\egroup
\vspace{0.5em}
\caption{\textit{Re-training Models}: comparison between the original network trained 
only with standard loss and a re-trained network using an adversarial loss term.}
\label{tab:re-train}
\end{table}

\begin{figure}[t]
    \centering
    \subfloat[\label{fig:base_2}]{%
      \includegraphics[width=0.22\textwidth]{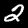}%
    } \hspace{0.75em}
    \subfloat[\label{fig:adv_2}]{%
      \includegraphics[width=0.22\textwidth]{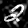}%
    } \hspace{0.75em}
    \subfloat[\label{fig:FourierTerms}]{%
      \includegraphics[width=0.44\textwidth]{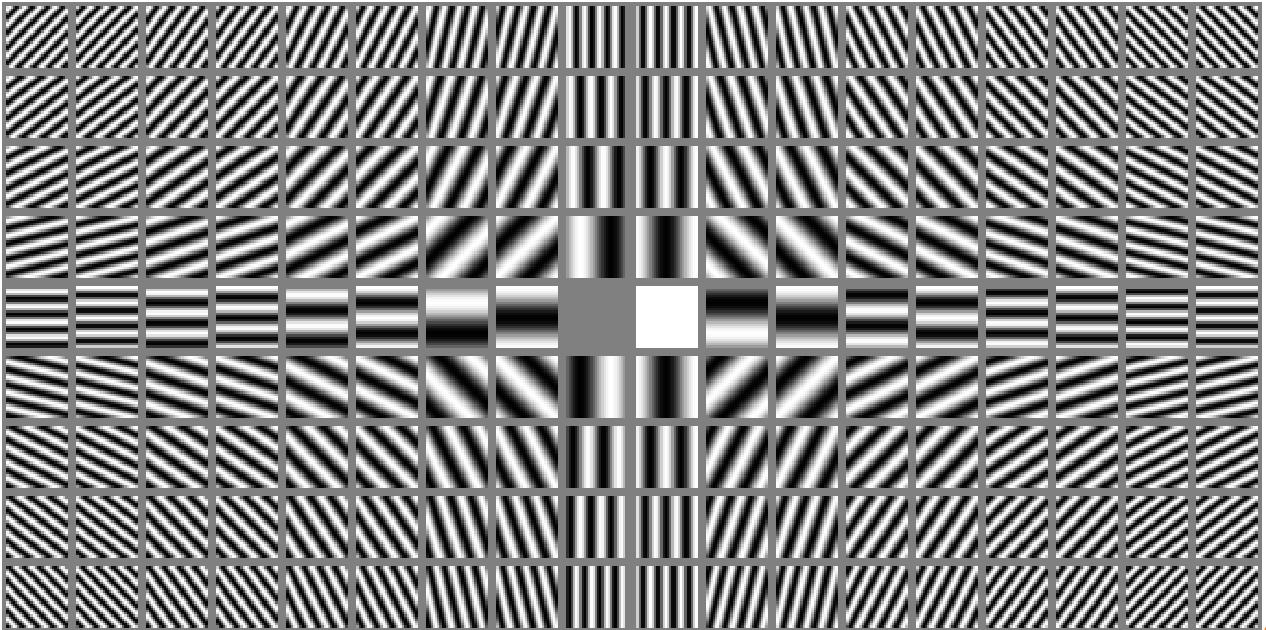}%
    }
    \caption{\textit{Fourier Attack}: example of a \texttt{Fourier} attack on an
    image of the MNIST dataset (a) an image that under regular training is correctly 
    identified as a \texttt{2} (b) an adversarial example identified using \framework
    and a \texttt{Fourier} attack which leads the network trained with a standard 
    loss function to identify as a \texttt{3}.
    (c) is a grid of thumbnails of the available Fourier terms that were added to (a).
    }
    \label{fig:fourier-example}
\end{figure}

\subsection{Custom Robustness Properties: Case Study}
As described in Section~\ref{sec:properties}, \framework includes a variety of 
built-in robustness properties on 2D images for \textsc{HybridZonotope}. 
In this section, we showcase the identification of adversarial examples based 
on the \texttt{Fourier} property.
Figure~\ref{fig:fourier-example} exemplifies an attack 
on a regularly trained network following the architecture of \texttt{ConvMed} 
(see \cite{pmlr-v80-mirman18b}).
The Fourier robustness property is motivated by the observation that a typical adversarial attack will include high frequency components which may be filtered away or rendered irrelevant by the variability in the real-world input image. It is interesting to ask whether adversarial examples exist only consisting of frequencies at roughly the scale of the original image.

In this scenario, the network correctly identifies the Figure~\ref{fig:base_2} 
as a \texttt{2}, yet is stumped by the adversarially generated image of
Figure~\ref{fig:adv_2} (using the Fourier terms presented in Figure~\ref{fig:FourierTerms}), mistakenly identifying it as a \texttt{3}. 
After training with \hz with $\lambda=0.1$ and 
$\epsilon=0.01$ for $200$ epochs for this robustness property, the model correctly identifies this specific example
as a \texttt{2}.

\section{Related Work}\label{s:related}
We consider three main areas of related work: early, \emph{heuristic} approaches to
training more robust networks; \emph{formal verification tools} that typically
operate on fully trained networks; and several other representative
\emph{robust training} approaches.
\paragraph{Heuristic approaches.}
Early art in adversarial robustness in the deep
learning community broadly tackled the problem with heuristic
techniques: with architecture and training scheme 
modifications~\cite{GuR14,ZhengStability,parseval,theoretical-robust}.
These techniques have shown  quite impressive results, and real
progress has been made to training more inherently robust
networks. 
However, it has been shown that these 
networks often remain susceptible to 
simple attacks~\cite{papernot-limitations,universal-perturbations}. 
This game of cat and mouse has led to ever more sophisticated
attack and defense,
\eg~\cite{GoodfellowExplainHarness,EnsembleAdversary,DBLP:journals/tnn/YuanHZL19,HuangLearningWith,MadryResistant,DongMomentum,DBLP:journals/access/AkhtarM18}. 
In terms of their usability, however, many of these early 
approaches are comparable to ours. A lot of the 
techniques involve modifications to the training scheme rather 
than the network architecture itself. 
As has been shown, this is broadly similar to how robust 
training can be applied within our framework.
The main limitation of these approaches is that they do not
provide guarantees for robustness and, 
as ever, the bad guys
tend to be one step ahead. 

\paragraph{Formal verification}
Formal verification techniques provide 
guarantees on the robustness of a \ac{dnn} at individual
data-points~\cite{marabou,reluplex,corina,DBLP:conf/kr/AkitundeLMP18,safetyverification},
and, in at least one case for a small single-layer network,
across the entirety of the input 
space~\cite{certified-adversarial}. 
Most of the work in this area focuses on \ac{smt}, reachability or
optimization-based approaches to provide sound and complete
guarantees on a per-example basis~\cite{nn-verification-survey}.
It has been shown that many of these techniques can be
viewed as flavours of a unified Branch-and-Bound framework~\cite{rudy-unified}.
Through this lens, one can see the scalability challenges as 
an artifact of the combinatorial branching associated with 
piecewise-linear activation functions such as ReLUs.
Similar to our framework, these formal verification tools require no modification 
to an existing codebase. In practice, most of the tooling is limited to a small 
subset of \ac{dnn} activations and layer types (e.g.,  convolutional networks 
are often not supported) limiting their utility in practice. 
Furthermore, the intractability of these
approaches, as detailed in~\cite{nn-verification-survey}, detracts their
use in many of the larger networks we study in this paper. Lastly,
as has been previously noted, these approaches do not offer systematic
improvements at scale, i.e., the verification or falsification of each point
needs to be considered iteratively in the training process.
\paragraph{Verifiably Robust Training}
Our work falls within a \emph{verifiably robust training} approach. 
We omit any theoretical comparison of the approaches, which is well 
described in~\cite{pmlr-v80-mirman18b}.
Our system is most closely similar to DiffAI~\cite{pmlr-v80-mirman18b}. 
However, it distinguishes itself in the way the abstract transformers
are generated from existing models: DiffAI requires that the user
specify their model using specialized classes.
This makes their library difficult to use with 
pre-existing models, since it requires rewriting the models 
to fit within the DiffAI framework.
Our framework, on the other hand, can take an existing TensorFlow
graph~\cite{abadi2016tensorflow} representing a model and
transform it automatically without having to rewrite any model
code, as shown in Section~\ref{s:system}.
This makes it more practical to use within an existing pipeline,
as it decouples the maintenance of the model from the verified
robustness procedure, allowing for faster development and testing. We observe that \framework
achieves similar performance to DiffAI and in a similar
total training times to those reported in~\cite{pmlr-v80-mirman18b} for the
same GPU configuration (Nvidia GeForce GTX 1080 Ti).

In~\cite{wang2018mixtrain}, the introduced framework, MixTrain, reaches better accuracy and a higher percentage of verifiably-safe
examples when compared to~\cite{pmlr-v80-mirman18b}. Similarly, comparing the
results presented in~\cite{wang2018mixtrain} with ours, we conclude that
MixTrain outperforms the ones obtained in Section~\ref{s:experiments}. However,
it should be noted that some of techniques that MixTrain uses  to achieve this improvement can be replicated easily when using our framework. For example, while in
Listing~\ref{lst:training_code} we defined the loss function as in
\cite{pmlr-v80-mirman18b} for the sake of simplicity, our framework allows
for flexible definitions, including the dynamic loss function defined per epoch in
\cite{wang2018mixtrain}. 

Other works in this area involve convex relaxation techniques such as the
ones presented in~\cite{salman2019convex}, or dual optimization techniques as
in~\cite{wong2018scaling}. In terms of accuracy and adversarial robustness,
further studies need to be carried out to compare our work to
\cite{salman2019convex} and~\cite{wong2018scaling}. Despite this, the
implementation of both~\cite{salman2019convex} and~\cite{wong2018scaling}
requires the re-writing of the models to adapt to the method's requirements, which,
as in the case of~\cite{pmlr-v80-mirman18b}, constitutes a set back to integration
efforts in production software stacks.

\section{Conclusion and Future Work}\label{s:conclusion}
In this paper we introduce \framework, a novel framework for verifiable robust training that 
can be used directly on existing codebases and requires minimal code changes.
We believe that this is the first practical framework for robust 
training that supports the vast majority of operations required for most large-scale models.
Our work further contributes to the community with the 
introduction of new abstract transformers, novel 
formal robustness properties, and a framework for adding user-defined properties to robust training.
We plan to build upon this framework in several directions:
\begin{compactitem}
    \item We wish to investigate more natural training
    schemes that, for example, use the robust loss more
    effectively  and  adapt the robustness property through
    the training cycle. 
    Similarly, we plan to explore how we could provide
    features such as the stochastic robust approximation techniques from
    \cite{wang2018mixtrain} for better performance.
    \item We also want to perform a theoretical study of abstract domains and training techniques that scale 
    better with larger \ac{dnn} widths and lengths; a fundamental problem of most of the methods presented in Section~\ref{s:related}~\cite{nn-verification-survey}. 
    It can be seen in Table~\ref{tab:diffai-comp}, for example, that {\hz}s did not perform as well as would be 
    expected on larger networks, and an in-depth analysis could help shed some light on the cause of this phenomenon.
    \item We wish to introduce an API for users to easily add and test their own op transformers, so that the framework can easily be extended to work on model code with currently unsupported ops. Currently supported ops may be viewed in Appendix~\ref{app:tf-ops}.
    \item Finally, we would like to conduct a comprehensive ablation study that includes many of the alternatives mentioned in Section~\ref{s:related} to further understand the comparative performance of our framework.
\end{compactitem}

\newpage
\bibliographystyle{splncs04}\bibliography{nfm-bib}
\newpage
\begin{subappendices}
\renewcommand{\thesection}{\Alph{section}}%

\section{Implemented TensorFlow Operations and Keras Layers}
\label{app:tf-ops}
Table~\ref{tab:tf-ops} lists the currently implemented TensorFlow operations in \framework, while Table~\ref{tab:keras-layers} shows the implemented Keras layers. Other Keras layers might be supported depending on the implementation in terms of TensorFlow operations.

\begin{table}[ht]
\centering
\bgroup
\def\arraystretch{1.3}
\begin{tabularx}{\textwidth}{*1{>{\arraybackslash}X}@{}|*2{>{\centering\arraybackslash}X}@{}}
\textbf{Operation Type} & \textsc{Box} & \textsc{HybridZonotope} \\ \hline
\tt{Abs} & & \checkmark \\ \hline
\tt{Add} & \checkmark & \checkmark \\ \hline
\tt{BiasAdd} & \checkmark & \checkmark \\ \hline
\tt{ConcatV2} & \checkmark & \checkmark \hspace{0.25cm}(only between HZ and tf.Tensor) \\ \hline
\tt{Conv2D} & \checkmark & \checkmark \hspace{0.25cm}(the second input should be tf.Tensor) \\ \hline
\tt{Exp} & & \checkmark \\ \hline
\tt{GreaterEqual} & \checkmark & \checkmark \\ \hline
\tt{Log} & & \checkmark \\ \hline
\tt{Log1p} & & \checkmark \\ \hline
\tt{MatMul} & \checkmark & \checkmark \hspace{0.25cm}(only first input HZ) \\ \hline
\tt{Maximum} & & \checkmark \\ \hline
\tt{MaxPool} & \checkmark & \checkmark \hspace{0.25cm}(only for `keras.MaxPool2D(2)) \\ \hline
\tt{Mean} & & \checkmark \\ \hline
\tt{Minimum} & & \checkmark \\ \hline
\tt{Mul} & \checkmark & \checkmark \\ \hline
\tt{Neg} & \checkmark & \checkmark \\ \hline
\tt{OnesLike} & \checkmark & \checkmark \\ \hline
\tt{Pack} & \checkmark & \\ \hline
\tt{RealDiv} & \checkmark & \checkmark \\ \hline
\tt{Relu} & \checkmark & \checkmark \\ \hline
\tt{Reshape} & \checkmark & \checkmark \\ \hline
\tt{Select} & & \checkmark \hspace{0.25cm}(first input not HZ) \\ \hline
\tt{Shape} & \checkmark & \checkmark \\ \hline
\tt{Sigmoid} & \checkmark & \checkmark \\ \hline
\tt{Softmax} & \checkmark & \checkmark \\ \hline
\tt{StridedSlice} & \checkmark & \checkmark \\ \hline
\tt{Sub} & \checkmark & \checkmark \\ \hline
\tt{Sum} & & \checkmark \\ \hline
\tt{Transpose} & \checkmark & \checkmark \\ \hline
\tt{ZerosLike} & \checkmark & \checkmark \\ \hline
\end{tabularx}
\egroup
\vspace{0.5em}
\caption{TensorFlow operations implemented in \framework}
\label{tab:tf-ops}
\end{table}

\begin{table}[ht]
\centering
\bgroup
\def\arraystretch{1.3}
\begin{tabularx}{\textwidth}{*1{>{\arraybackslash}X}@{}|*2{>{\centering\arraybackslash}X}@{}}
\textbf{Layer} & \textsc{Box} & \textsc{HybridZonotope} \\ \hline
\tt{Concatenate} & \checkmark & \checkmark \\ \hline
\tt{Conv2D} & \checkmark & \checkmark \\ \hline
\tt{Dense('relu')} & \checkmark & \checkmark \\ \hline
\tt{Dense('sigmoid')} & \checkmark & \checkmark \\ \hline
\tt{Dense('softmax')} & \checkmark & \checkmark \\ \hline
\tt{Flatten} & \checkmark & \checkmark \\ \hline
\tt{MaxPooling2D} & \checkmark & \checkmark \\ \hline
\end{tabularx}
\egroup
\vspace{0.5em}
\caption{Keras layers supported by \framework out of the box}
\label{tab:keras-layers}
\end{table}

\section{Network architectures}
\label{app:networks}

We follow the design of~\cite{pmlr-v80-mirman18b}. For convolutional layers $c\times w\times h\; [s]$ is for channels, kernel width, kernel height and stride, respectively.
\paragraph{FFNN} Five fully-connected layers, 100-node each, 
with ReLU.
\paragraph{ConvSmall} Two convolutional layers with no padding ($16\times4\times4$~[2], $32\times4\times4$~[2]), followed by a 100-node fully-connected layer.
\paragraph{ConvMed} Two convolutional layers with padding of 1 ($16\times4\times4$~[2], $32\times4\times4$~[2]), followed by a 100-node fully-connected layer.
\paragraph{ConvBig} Four convolutional layers with padding of 1 (
$32 \times 3 \times 3$~[1], $32 \times 4 \times 4$~[2], $64 \times 3 \times 3$~[1], $64 \times 4 \times 4$~[2]), followed by a 512-node fully-connected layer, ReLU, and a 512-node fully-connected layer.
\paragraph{ConvSuper} Four convolutional layers with no padding ( $32 \times 3 \times 3$~[1] , $32 \times 4 \times 4$~[1], $64 \times 3 \times 3$~[1], $64 \times 4 \times 4$~[1]), followed by a 512-node fully-connected layer, ReLU, and a 512-node fully-connected layer.
\paragraph{Skip} A concatenation of two covolutional networks followed by ReLU, 200-node fully-connected network, and ReLU. The two networks are:
\begin{itemize}
    \item Three convolutional layers ($16 \times 3 \times 3$~[1], $16 \times 3 \times 3$~[1], $32 \times 3 \times 3$~[1]), followed by a  200-node fully-connected layer
    \item Two convolutional layers ($32 \times 4 \times 4$~[1], $32 \times 4 \times 4$~[1]) followed by a 200-node fully-connected layer.
\end{itemize} 

\end{subappendices}

\end{document}